\documentclass[11pt,a4paper]{article}
\usepackage[hyperref]{naaclhlt2019}
\usepackage{times}
\usepackage{latexsym}
\usepackage{booktabs}
\usepackage[nolist]{acronym}
\usepackage{url}
\usepackage{todonotes}
\usepackage{array}

\usepackage{tabularx}
\usepackage{dcolumn}
\usepackage{longtable}
\usepackage{supertabular}
\usepackage[inline]{enumitem}
\usepackage{xspace}
\usepackage{listings}
\usepackage{fancyvrb}
\usepackage{comment}
\usepackage{wrapfig}  
\usepackage{caption}
\usepackage{subcaption}
\usepackage{longtable}
\usepackage{multirow}
\usepackage{colortbl}
\usepackage[T1]{fontenc}
\usepackage{tikz}
\usepackage{footnote}
\usepackage{amssymb}
\usepackage[ruled,vlined,linesnumbered]{algorithm2e}
\usepackage{url}
\usepackage{mathtools}
\usepackage{verbatim}
\usepackage{lipsum}
\usepackage{listingsutf8}
\usepackage{lscape}
\usepackage{pifont}
\usepackage{numprint}
\usepackage{adjustbox}
\usepackage{multirow}
\usepackage{multicol}
\usepackage{float}

\usepackage[scaled=.7]{beramono}

\newtheorem{definition}{Definition}

\newcolumntype{R}[2]{%
    >{\adjustbox{angle=#1,lap=\width-(#2)}\bgroup}%
    l%
    <{\egroup}%
}

\newcommand{\tablefont}[1]{\fontsize{3mm}{3.2mm}\selectfont}

\aclfinalcopy 

\title{Augmenting Neural Machine Translation with Knowledge Graphs}

\author{Diego Moussallem$^{1}$ \, Mihael Ar\v{c}an$^{2}$ \, Axel-Cyrille Ngonga Ngomo$^{1}$ \, Paul Buitelaar$^{2}$ \\
         $^{1}$Data Science Group, University of Paderborn, Germany \\
         $^{2}$Insight Centre for Data Analytics, Data Science Institute \\ National University of Ireland Galway \\ 
        {{\tt \{first.lastname\}@upb.de}} \\ {\tt \{firstname.lastname\}@insight-centre.org}}

\begin{acronym}[UML]
	\acro{AOS}{Agricultural Ontology Services}
	\acro{AGRIS}{Agricultural Science and Technology}
	\acro{API}{Application Programming Interface}
	\acro{AOS}{Agricultural Ontology Services}
	\acro{AGRIS}{Agricultural Science and Technology}
	\acro{API}{Application Programming Interface}
	\acro{A2KB}{Annotation to Knowledge Base}
	\acro{BPSO}{Binary Particle-Swarm Optimization}
	\acro{BPMLOD}{Best Practices for Multilingual Linked Open Data}
	\acro{BPSO}{Binary Particle-Swarm Optimization}
	\acro{BPMLOD}{Best Practices for Multilingual Linked Open Data}
	\acro{BFS}{Breadth-First-Search}
	\acro{BPE}{Byte Pair Encoding}
	\acro{BoW}{Bag-of-Words}
	\acro{CBD}{Concise Bounded Description}
	\acro{COG}{Content Oriented Guidelines}
	\acro{CSV}{Comma-Separated Values}
	\acro{CBMT}{Corpus-Based Machine Translation}
	\acro{CLIR}{Cross-Language Information Retrieval}
	\acro{DPSO}{Deterministic Particle-Swarm Optimization}
	\acro{DALY}{Disability Adjusted Life Year}

	\acro{ER}{Entity Resolution}
	\acro{EM}{Expectation Maximization}
	\acro{EBMT}{Example-Based Machine Translation}
	\acro{EBNF}{Extended Backus--Naur Form}
	\acro{EL}{Entity Linking}
	\acro{FAO}{Food and Agriculture Organization of the United Nations}
	\acro{GIS}{Geographic Information Systems}
	\acro{GHO}{Global Health Observatory}
	\acro{GRU}{Gated recurrent unit}
	\acro{HDI}{Human Development Index}
	\acro{ICT}{Information and communication technologies}
	\acro{IFRS}{International Financial Reporting Standards}
	\acro{ICD}{International Classification of Diseases}
	\acro{IT}{Information Technology}
    \acro{KB}{Knowledge Base}
    \acro{KG}{Knowledge Graph}
    \acro{KGE}{Knowledge Graph Embeddings}
    \acro{KBSE}{Knowledge Base Semantic Embedding}
	\acro{LR}  {Language Resource}
	\acro{LD}  {Linked Data}
	\acro{LLOD}  {Linguistic Linked Open Data}
	\acro{LIMES}{LInk discovery framework for MEtric Spaces}
	\acro{LS}  {Link Specifications}
	\acro{LDIF}{Linked Data Integration Framework}
	\acro{LGD} {LinkedGeoData}
	\acro{LOD} {Linked Open Data}
	\acro{LSTM}{Long Short-Term Memories}
	\acro{MSE}{Mean Squared Error}
	\acro{MWE}{Multiword Expressions}
	\acro{MT}{Machine Translation}
	\acro{ML}{Machine Learning}
	\acro{MR}{Machine Reading}
	\acro{NIF}{Natural Language Processing Interchange Format}
	\acro{NIF4OGGD}{NLP Interchange Format for Open German Governmental Data}
	\acro{NLP}{Natural Language Processing}
	\acro{NER}{Named Entity Recognition}
	\acro{NMT}{Neural Machine Translation}
	\acro{NN}{Neural Network}
	\acro{NLG}{Natural Language Generation}
	\acro{NED}{Named Entity Disambiguation}
	\acro{NERD}{Named Entity Recognition and Disambiguation}
	\acro{NL}{Natural Language}
	\acro{NIF}{NLP Interchange Format}
	\acro{NIF4OGGD}{NLP Interchange Format for Open German Governmental Data}
	\acro{NLP}{Natural Language Processing}
	\acro{NER}{Named Entity Recognition}
	\acro{NEL}{Named Entity Linking}
	\acro{NE}{Named Entity}
	\acro{NN}{Neural Network}
	\acro{NLI}{Natural Language Inference}
	\acro{OSM}{OpenStreetMap}
	\acro{OWL}{Web Ontology Language}
	\acro{OOV}{out-of-vocabulary}
	\acro{PFM}{Pseudo-F-Measures}
	\acro{PSO}{Particle-Swarm Optimization}
	\acro{PBSMT}{Phrase-Based Statistical Machine Translation}
	
	\acro{QA}{Question Answering}
	\acro{RDF}{Resource Description Framework}
	\acro{RBMT}{Rule-Based Machine Translation}
	\acro{RNN}{Recurrent Neural Network}
	\acro{ReLU}{rectified linear unit}
	\acro{SKOS}{Simple Knowledge Organization System}
	\acro{SPARQL}{SPARQL Protocol and RDF Query Language}
	\acro{SRL}{Statistical Relational Learning}
	\acro{SWT}{Semantic Web Technologies}
	\acro{SW}{Semantic Web}
	\acro{SMT}{Statistical Machine Translation}
	\acro{SWMT}{Semantic Web Machine Translation}
	\acro{SKOS}{Simple Knowledge Organization System}
	\acro{SPARQL}{SPARQL Protocol and RDF Query Language}
	\acro{SRL}{Statistical Relational Learning}
	\acro{SF}{surface forms}

    \acro{TBMT} {Transfer-Based Machine Translation}
	\acro{UML}{Unified Modeling Language}
	\acro{USL}{Ukrainian Sign Language}
	\acro{URI}{Uniform Resource Identifier}
	\acro{WHO}{World Health Organization}
	\acro{WKT}{Well-Known Text}
	\acro{W3C}{World Wide Web Consortium}
	\acro{WSD}{Word Sense Disambiguation}
    \acro{XML}{Extensible Markup Language}
	\acro{YPLL}{Years of Potential Life Lost}

\end{acronym}  

\begin{document}
\maketitle
\begin{abstract}

While neural networks have been used extensively to make substantial progress in the machine translation task, they are known for being heavily dependent on the availability of large amounts of training data. Recent efforts have tried to alleviate the data sparsity problem by augmenting the training data using different strategies, such as back-translation. Along with the data scarcity, the out-of-vocabulary words, mostly entities and terminological expressions, pose a difficult challenge to Neural Machine Translation systems. In this paper, we hypothesize that knowledge graphs enhance the semantic feature extraction of neural models, thus optimizing the translation of entities and terminological expressions in texts and consequently leading to a better translation quality. 
We hence investigate two different strategies for incorporating knowledge graphs into neural models without modifying the neural network architectures. We also examine the effectiveness of our augmentation method to recurrent and non-recurrent (self-attentional) neural architectures. 
Our knowledge graph augmented neural translation model, dubbed KG-NMT, achieves significant and consistent improvements of +3 {\sc BLEU}, {\sc METEOR} and {\sc chrF3} on average on the \textit{newstest} datasets between 2014 and 2018 for WMT English-German translation task.

\end{abstract}

\section{Introduction}

\ac{NN} models have shown significant improvements in translation generation and have been widely adopted given their sustained improvements over the previous state-of-the-art \ac{PBSMT} approaches \cite{koehn2007moses}. A number of \ac{NN} architectures have therefore been proposed in the recent past, ranging from recurrent~\cite{bahdanau2014neural,sutskever2014sequence} to self-attentional networks~\cite{vaswani2017attention}. 
However, a major drawback of \ac{NMT} models is that they need large amounts of training data to return adequate results and have a limited vocabulary size due to their computational complexity~\cite{luong2016achieving}. The data sparsity problem in \ac{MT}, which is mostly caused by a lack of training data, manifests itself in particular in the poor translation of rare and \ac{OOV} words, e.g., entities or terminological expressions rarely or never seen in the training phase. 
Previous work has attempted to deal with the data sparsity problem by introducing character-based models~\cite{luong2016achieving} or \ac{BPE} algorithms~\cite{sennrich2015neural}. Additionally, different strategies were devised for overcoming the lack of training data, such as back-translation~\cite{sennrich2016improving}.

Despite the significant advancement of previous work in \ac{NMT}, translating entities and terminological expressions remains a challenge~\cite{koehn2017six}. 
Entities may be subsumed in two groups, i.e., proper nouns and common nouns. 
Proper nouns are also known as \ac{NE} and correspond to the name of persons, organizations or locations, e.g., \textit{Canada}. Common nouns 
describe classes of object, e.g., \textit{spoon} or \textit{cancer}. Both types of entities are found in a \ac{KG}, in which they are described within triples~\cite{auer2007dbpedia,vrandevcic2014wikidata}. 
Each triple consists of a subject---often an entity---, a relation---often called property---and an object---often an entity or a literal, e.g., a string or a value with a unit---. For example, \texttt{<NAACL, areaServed, North\_America>},\footnote{\url{http://dbpedia.org/resource/NAACL}} means in natural language that ``NAACL takes place in North America''.\footnote{In this paper, we use \ac{KG} and KB interchangeably.} Recent work has exploited the contribution of \ac{KG}s to improve distinct \ac{NLP} tasks such as \ac{NLI}~\cite{annervaz2018},
\ac{QA}~\cite{Sorokincoling2018,sun2018open} and \ac{MR}~\cite{yang2017leveraging} successfully. Additionally, the benefits of incorporating type information on entities---e.g., \ac{NE}-tags such as \texttt{PERSON}, \texttt{LOCATION} or \texttt{ORGANIZATION}---into \ac{NMT} by relying on \ac{NER} systems have been shown in previous works~\cite{ugawa2018neural,li2018named}. However, none of these have exploited the combination of \ac{EL} with \ac{KG}s in \ac{NMT} systems. 

The goal of \ac{EL} is to disambiguate and link a given \ac{NE} contained in a text to a corresponding entity---also called a resource---in a reference \ac{KB} \cite{moussallem2017mag}. If the  reference \ac{KB} is bilingual, then the links generated by \ac{EL} can be used to retrieve the translation of entities found in the text. In this work, we aim to use \ac{EL} to improve the results of \ac{NMT} approaches. We build upon recent works, which have devised \ac{KGE} approaches~\cite{TransE/bordes2013translating}, i.e., approaches that embed \ac{KG}s into continuous vector spaces. Since neural models learn translations in a continuous vector space, we hypothesize that a given \ac{KG}, once converted to embeddings, can be used along with \ac{EL} to improve \ac{NMT} models. Our results suggest that with this proposed methodology, we are capable of enhancing the semantic feature extraction of neural models for gathering the correct translation of entities and consequently improving the translation quality of the text.

We devised two strategies to implement the insight stated above. In our first strategy, we began by annotating bilingual training data with a multilingual \ac{EL} system using a reference \ac{KB}. Then, we map the entities and relationships contained in the reference \ac{KB} to a continuous vector space using a \ac{KGE} technique. Afterwards, we concatenate the \ac{KGE} to the internal \ac{NMT} embeddings, thus augmenting the embedding layer of \ac{NMT} training. Given that \ac{EL} can be time-consuming when faced with large training corpora, we skip the \ac{EL} task in our second strategy and we semantically enrich the \ac{KGE} by using the referring expressions of entities, also known as labels to initialize the vector values at the embedding layer. Differently from \citet{venugopalan2016improving}, we maximize the vector values of entities found in the bilingual corpora with the values of entities' labels from the \ac{KGE}. We perform an extensive automatic and manual evaluation in order to analyze our hypothesis. Among others, we examine the effectiveness of our augmentation method when combined with recurrent and non-recurrent (self-attentional) neural architectures, dubbed RNN and Transformer respectively. Our \ac{KG}-augmented neural translation model, named KG-NMT, achieves significant and consistent improvements of +3 {\sc BLEU}, {\sc METEOR} and {\sc chrF3} on average on the WMT \textit{newstest} datasets between 2014 and 2018 for the English-German translation task, using a small set of two million parallel sentences. To the best of our knowledge, no previous work has investigated the augmentation of \ac{NMT} by using \ac{KG}s without affecting the \ac{NN} architecture. Hence, the main contribution of this paper lies in the investigation of two different strategies for integrating \ac{KGE} into neural translation models to maximize the probability score of the translation of entities. Moreover, we show that we can enhance the translation quality of \ac{NMT} systems by incorporating \ac{KGE} into the training phase.\footnote{Our data and models will be made publicly available.}

\section{Related Work}

\vspace{-2mm}
\paragraph{NMT Augmentation.}

Different methods have been suggested to overcome the limitations of \ac{NMT} vocabulary size. \citet{luong2016achieving} implemented a hybrid solution, which combines word and character models in order to achieve an open vocabulary \ac{NMT} system. 
Similarly, \citet{sennrich2015neural} introduced \ac{BPE}, which is a form of data compression that iteratively replaces the most frequent pair of bytes in a sequence with a single, unused byte. Additionally, the use of monolingual data for data augmentation has gained considerable attention as it is not supposed to alter the \ac{NN} architecture while demonstrating consistent results. \citet{sennrich2016improving} explored two methods using monolingual data during the training of an \ac{NMT} system. They used dummy source sentences and relied on an automatic back-translation of the monolingual data using different \ac{NMT} systems. Moreover, past work exploited the use of monolingual data to augment \ac{NMT} systems in distinct \ac{NN} architectures. \citet{hoang2018iterative} presented an iterative back-translation method, which generates increasingly synthetic parallel data from monolingual data while training a given \ac{NMT} system. Also, \citet{edunov2018understanding} attempted to understand the effectiveness of back-translation in a large scale scenario by using different strategies on hundreds of millions of monolingual sentences. Recently, approaches other than back-translation for data augmentation were introduced. For example, \citet{wang2018switchout} proposed a method of randomly replacing words in both the source sentence and the target sentence with other random words from their corresponding vocabularies. 

\vspace{-2mm}
\paragraph{External Structured Knowledge in \ac{MT}.}
According to a recent survey~\cite{moussallem2018machine}, the idea of using a structured \ac{KB} in \ac{MT} systems started with the work of \citet{knight1994building}. Still, only a few researchers have designed different strategies for benefiting of  structured knowledge in \ac{MT} architectures~\cite{mccrae2013mining,arcan2015knowledge,simovtowards}. Recently, the idea of using \ac{KG} into \ac{MT} systems 
has gained renewed attention. \citet{jinhua2016} created an approach to address the problem of \ac{OOV} words by using BabelNet~\cite{navigli2012babelnet}. Their approach applies different methods of using BabelNet. In sum, they create additional training data and also apply a post-editing technique which replaces the \ac{OOV} words while querying BabelNet.
\citet{shi2016knowledge} have recently built a semantic embedding model reliant upon a specific \ac{KB} to be used in \ac{NMT} systems. The model relies on semantic embeddings to encode the key information contained in words so as to translate the meaning of sentences correctly. 
\vspace{-2mm}
\paragraph{Named Entities in \ac{NMT}.}

Only a few works have investigated the \ac{NE} translation issue in \ac{NMT}. Some researchers worked on models specific to this problem, while others incorporated external information as features within \ac{NMT} models. \citet{li2016neural} and \citet{wang2017sogou} rely on an \ac{NER} tool to identify and align the \ac{NE} pairs within the source and target sentences. Afterwards, the \ac{NE} pairs are replaced with their corresponding \ac{NE}-tags to train the model. In the translation phase, the targeted \ac{NE} tags are then substituted with the original entities by a separate \ac{NE} translation model or a bilingual \ac{NE} dictionary. \citet{ugawa2018neural} used a similar architecture but included one more layer in the encoder to encode the \ac{NE}-tags expressed as chunk tags at each time step. The disadvantages of the methods above include \ac{NE} information loss and \ac{NE} alignment errors. To overcome these problems, \citet{li2018named} relied on an effective and simple method which added the \ac{NE}-tags as boundary information to the entities directly inserted by an \ac{NER} tool in the source sentence. It does not require either any separate model or external resource, and it therefore does not affect the \ac{NN} architecture while achieving good performance.
\vspace{-2mm}
\paragraph{Knowledge Graph Embeddings.}

According to \citet{annervaz2018}, we classify \ac{KGE} into two categories: (1) Structure-based, which encodes only entities and relations, (2) Semantically-enriched, which takes into account semantic information of entities, e.g., text, along with the entities and its relations. (1) According to ~\citet{wang2017knowledge} manifold approaches, where relationships are interpreted as displacements operating on the low-dimensional embeddings of the entities, have been implemented so far, such as TransE~\cite{TransE/bordes2013translating} and TransG~\cite{TransG/xiao2015transg}. However, \citet{joulin2017fast} showed recently that a simple \ac{BoW} based approach with the fastText algorithm~\cite{joulin2017bag} generates surprisingly good \ac{KGE} while achieving the state-of-the-art results. (2) \citet{KGtext/wang2014knowledge} proposed a technique of learning to embed structured and unstructured data (such as text) jointly in an effort to augment the prediction models. Additionally, \citet{zhong2015aligning} introduced an alignment of entities and word embeddings considering the description of entities. More work on agglutinating the semantics with entities arose, such as SSP~\cite{xiao2017ssp} and DKRL~\cite{xie2016representation} as well as TKRL~\cite{xie2016representationtkrl}.

 \section{The KG-NMT Methodology}
KG-NMT is based on the observation that more than 150 billion facts referring to more than 3 billion entities are available in the form of \ac{KG} on the Web~\cite{mccrae2018lod}. Hence, the intuition behind our methodology is as follows: \emph{Given that \ac{KG}s describe real-world entities, we can use \ac{KG}s along with \ac{EL} to optimize the entries in the vector of entities and consequently to achieve a better translation quality of entities in text}. 
In the following, we give an overview of \ac{NMT} and \ac{KGE}. Afterwards, we present how we use \ac{EL} and \ac{KGE} to augment \ac{NMT} models. Throughout the description of our methodology and our experiments, we used DBpedia~\cite{auer2007dbpedia} as reference \ac{KB}. 

\subsection{Background}

\subsubsection{Neural Machine Translation}
We use two different \ac{NMT} architectures, the \ac{RNN} and Transformer-based models. Both consist of an encoder and a decoder, i.e., a two-tier architecture where the encoder reads an input sequence $x=(x_1,...,x_n)$ and the decoder predicts a target sequence $y=(y_1,...,y_n)$. The encoder and decoder interact via a soft-attention mechanism \cite{bahdanau2014neural,LuongAttention2015}, which comprises of one or multiple attention layers. We follow the notations from \citet{tang2018self} in the subsequent sections: 
$h^{l}_{i}$ corresponds to the hidden state at step $i$ of layer $l$. $h^{l}_{i-1}$ represents the hidden state at the previous step of layer $l$ while $h^{l-1}_{i}$ means the hidden state at $i$ of $l-1$ layer. $E \in \mathbb{R}^{m\times K_x}$ is a word embedding matrix, $W \in \mathbb{R}^{n\times m}$, $U \in \mathbb{R}^{n\times n}$ are weight matrices, with $m$ being the word embedding size and $n$ the number of hidden units. $K_x$ is the vocabulary size of the source language. Thus, $E_{x_{i}}$ refers to the embedding of $x_{i}$, and $e_{pos,i}$ indicates the positional embedding at position $i$. 
\vspace{-2mm}
\paragraph{RNN-based NMT.}
  \label{ssub:rnn_based_nmt}

In \ac{RNN} models, networks change as new inputs (previous hidden state and the token in the line) come in, and each state is directly connected to the previous state only. Therefore, the path length of any two tokens with a distance of $n$ in RNNs is exactly $n$. Its architecture enables adding more layers, whereby two adjoining layers are usually connected with residual connections in deeper configurations. Equation \ref{eq:rnn-enc-hidden} displays $h^{l}_{i}$, where 
$f_{rnn}$ is usually a function based on \ac{GRU}~\cite{cho2014learning} or \ac{LSTM}~\cite{hochreiter1997long}. The first layer is then represented as $h^{0}_{i} = f_{rnn}(WE_{x_{i}}, U h^{0}_{i-1})$. Additionally, the initial state of the decoder is commonly initialized with the average of the hidden states or the last hidden state of the encoder. 

\vspace{-2mm}
\begin{equation} \label{eq:rnn-enc-hidden}
h^{l}_{i} = h^{l-1}_{i} + f_{rnn}(h^{l-1}_{i}, h^{l}_{i-1})
\end{equation} 
\vspace{-2mm}

\vspace{-2mm}
\paragraph{Transformer-based NMT.}
\label{ssub:transformer_based_nmt}

Transformer models rely deeply on self-attention networks. Each token is connected to any other token in the same sentence directly via self-attention. Thus, the path length between any two tokens is $1$. Additionally, these models rely on multi-head attention to feature attention networks, which are more complex in comparison to $1$-head attention mechanisms used in RNNs. In contrast to RNN, the positional information is also preserved in positional embeddings. Equation \ref{eq:self-attention} represents the hidden state $h^{l}_{i}$, which is calculated from all hidden states of the previous layer. $f$ represents a feed-forward network with the \ac{ReLU} as the activation function and layer normalization. The first layer is represented as $h^{0}_{i} = WE_{x_{i}} + e_{pos,i}$. Moreover, the decoder has a multi-head attention over the encoder hidden states.

\vspace{-2mm}
\begin{equation} \label{eq:self-attention}
h^{l}_{i} = h^{l-1}_{i} + f(\text{self-attention}(h^{l-1}_{i}))
\end{equation}

\vspace{-2mm}
\subsubsection{Knowledge Graph Embeddings}

The underlying concept of \ac{KGE} is that, in a given \ac{KB}, each subject $h$ or object $t$ entity can be associated as a point in a continuous vector space whereby its relation $r$ can be modelled as displacement vectors ($h + r = t$) while preserving the inherent structure of the \ac{KG}. In the methodology introduced by \citet{joulin2017fast}, named fastText, the model is based on \ac{BoW} representation which considers the subject $h$ and object $t$ entities along with its relation $r$ as a unique discrete token. Thus, fastText models the co-occurrences of entities and its relations with a linear classifier and standard cost functions. Hence, it allows theoretically creating either a Structure-based or Semantically-enriched \ac{KGE}. Therefore, we use fastText models in our experiments, represented by the following equation~\autoref{eq:fastText}. 

\vspace{-2mm}
\begin{equation}
\label{eq:fastText}
-\frac{1}{N} \sum_{n=1}^N y_n \log( f (WVz_n)),
\end{equation}
\vspace{-2mm}

The normalized \ac{BoW} of the~$x_n$~input set is represented as $z_n$,~$y_n$ as the label. $V$ is a matrix, which is used as a look-up table over the discrete tokens and a matrix~$W$ is used for the classifier.  The representations of the discrete tokens are averaged into \ac{BoW} representation, which is in turn fed to the linear classifier. $f$ is used to compute the probability distribution over the classes, and~$N$ input sets for discrete tokens. We denote the generated \ac{KGE} as $E'$.


\subsection{Methodology}
\label{sec:kg-nmt}

\begin{figure*}[htb]
\centering
    \includegraphics[width=0.8\textwidth]{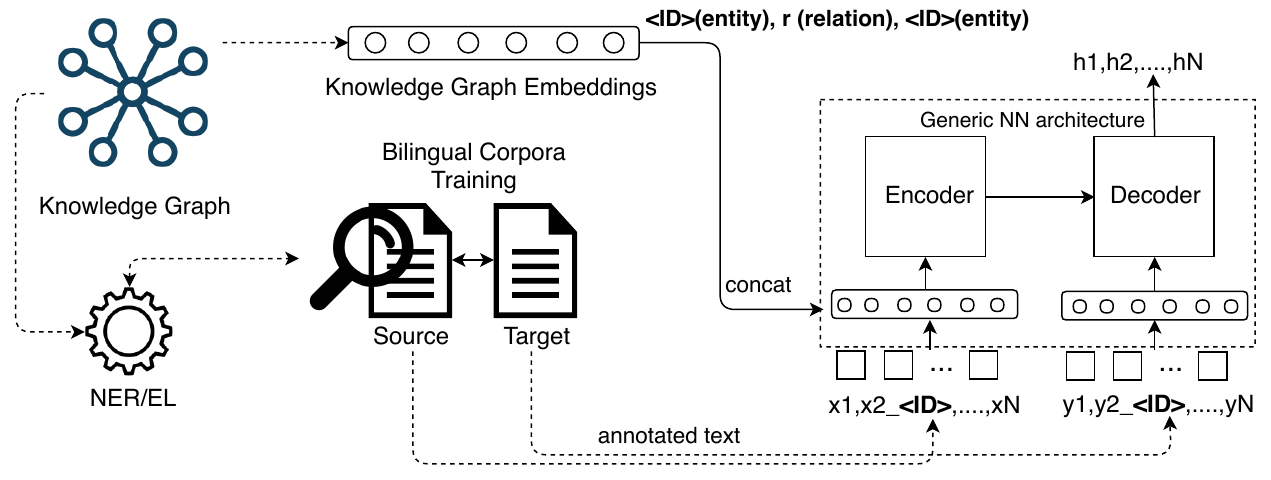}
    \vspace{-2mm}
    \caption{Overview of the KG-NMT methodology.}
    \vspace{-2mm}
    \label{fig:arch}
\end{figure*}

Recent work has successfully devised strategies for incorporating different kinds of knowledge into \ac{NMT} models, such as linguistic features \cite{sennrich2016linguistic} and \ac{NE}-tags~\cite{gu2016incorporating}.
Differently---but inspired by the above-mentioned approaches---instead of training a given \ac{NMT} model on a large amount of parallel data with the aim of improving the translation of entities, our idea relies on \ac{EL} solutions, which can disambiguate the entities found in a text to translate by mapping them to a corresponding node from a given reference \ac{KG}.
In turn, the \ac{KG} can support the learning of a neural translation models through its graph structure.
~\autoref{fig:arch} depicts the general idea of our methodology. Formally, we substantiate our methodology on the two following definitions of \ac{EL} and \ac{KB}.\footnote{We assume that all mentions can be linked to entities in the \ac{KB}}  

\begin{definition}
Entity Linking:
Let $\mathcal{E}$ be a set of entities from a \ac{KB} and $\mathcal{D}$ be a document containing potential mentions of entities \textbf{m} = $({m}_1,\dots,{m}_n)$. The goal of an \ac{EL} system is to generate an assignment $\mathcal{F}$ of mentions to entities with $\mathcal{F}(\textbf{m}) \in  (\mathcal{E)}^n$ for the document $\mathcal{D}$.
\end{definition}

\begin{definition}
\label{def:el}
Knowledge Base: We define \ac{KB} $K$ as a directed graph $G_K = (V, R)$ where the nodes $V$ are resources of $K$, the edges $R$ are properties of $K$ and $h,t\in V, (h,t) \in R \Leftrightarrow \exists r : (h, r, t) \mbox{ is a triple in }K$.
\end{definition}

We devised two strategies to instantiate our methodology. In the first training strategy, we link the \ac{NE}s in the source and target texts to a reference \ac{KB} using a given multilingual \ac{EL} system. We then incorporate the \ac{URI}s of entities along with the tokens akin to \citet{li2018named} with the \ac{NE}-tags. For example, the word \textit{Kiwi} can be annotated with \texttt{Ki\-wi|dbr\_Kiwi}\footnote{\url{http://dbpedia.org/resource/Kiwi}} or \texttt{dbr\_Kiwi\_(people)}, depending on the context. Similarly, the word \textit{cancer} can be annotated with \texttt{cancer|dbr\_Cancer},\footnote{\url{http://dbpedia.org/resource/Cancer}} and its translation can be found in the German part of the DBpedia \ac{KB} (\texttt{dbr\_Krebs\_(Medizin)}). After incorporating the URIs, we embed the reference \ac{KB}, DBpedia, using the fastText \ac{KGE} algorithm. Once the \ac{KGE} embeddings are created, we concatenate their vectors to the internal vectors of \ac{NMT} embeddings. The concatenation is possible as the annotations, i.e., URIs, are present in the texts and consequently in the vocabulary~\cite{speer2017conceptnet}. Formally, let the tokens from the source and target text be elements of a fixed vocabulary $\mathcal{D}$ which are used to train a given \ac{NMT} model, while the assignments $\mathcal{F}$ are the nodes $V$ within \ac{KB} $K$. The embeddings $E'$ of $K$ can be concatenated along with the internal embeddings of \ac{NMT} $E$ using a function $concat(E,E')$, thus resulting in a new vector $E_c$. With this modification the first layer of an RNN becomes $h^{0}_{i} = f_{rnn}(W{E_c}_{x_{i}}, Uh^{0}_{i-1})$. 

Although incorporating \ac{EL} as a feature into \ac{NMT} is interesting by itself, the annotation of entities in the training set and the post-editing can be resource-intensive. Additionally, one limitation of Structure-based \ac{KGE}s is that it can only work with word-based models since it is not possible to apply any segmentation model on entities and relations, since segmentation may force the algorithm to assign wrong vectors to the entities. For example, the entities \textit{dbr:Leipzig} and \textit{dbr:Leibniz} can be similar when considering sub-word units, however, the first is a location while the second is a person. Thus, they should not be regarded as similar. To overcome both limitations, we devised our second strategy which uses only Semantically-enriched \ac{KGE}s and skips the \ac{EL} part. Here, we enrich the Structure-based \ac{KGE} with referring expressions of the entities found in the \ac{KB}, thus decreasing the annotation effort. To generate the Semantically-enriched \ac{KGE}, we rely on a classifier in a supervised training implemented in fastText which assigns a label to a given entity. For example, we add to the triple, \texttt{<NAACL, areaServed, USA>} the following information, \texttt{<NAACL, label, North American Chapter of the Association for Computational Linguistics>}.\footnote{More than one label can be assigned to the entities.} By enriching the \ac{KGE}, it allows us to use the vectors to initialize the embedding layer's weights of the \ac{NMT} models similarly to \citet{neishi2017bag}, which used pre-trained monolingual embeddings. Furthermore, it also enables applying segmentation to the labels, which allows work with \ac{BPE} models. 
Commonly, the initialization of the embeddings layer is a function which assigns random values to the weight matrix $W$, whereas in our second strategy, the values from \ac{KGE} $E'$ matrix are used to assign constant values to matrix $W$ using a function $init(E')$.

 \section{Experimental Setup}

In our experiments, we used the multilingual \ac{EL} system introduced by \citet{moussallem2017mag} which is language and \ac{KB} agnostic. Also, it does not require any training and still has shown competitive results according to the benchmarking platform GERBIL~\cite{gerbil}. Different \ac{NN} architectures are complex to compare as they are susceptible to hyper-parameters. Therefore, the idea was to use a minimal reasonable configuration set in order to allow a fair analysis of the real \ac{KG} contributions. For our overall experiments, the \ac{RNN}-based models use a bi-directional 2-layer \ac{LSTM} encoder-decoder model with attention~\cite{bahdanau2014neural}. The training uses a batch size of 32 and the stochastic gradient descent with an initial learning rate of 0.0002. We set a word embeddings' size of 500, and hidden layers to size 500, dropout = 0.3 (naive). We use a maximum sentence length of 80, a vocabulary of 50 thousand words and a beam size of 5. All experiments were performed with OpenNMT~\cite{2017opennmt}. In addition, we encoded words using \ac{BPE}~\cite{sennrich2015neural} with 32,000 merge operations to achieve an open vocabulary. OpenNMT enables substitution of \ac{OOV} words with target words that have the highest attention weight according to their source words~\cite{LuongAttention2015} and when the words are not found, it uses a copy mechanism which copies the source words to the position of the not-found target word~\cite{gu2016incorporating}. Thus, we used all the options mentioned above to evaluate the performance of the translation quality. We trained the \ac{KGE}s with a vector dimension size of 500 with a window size of 50 by using 12 threads with hierarchical softmax. In addition, to Semantically-enriched \ac{KGE} we added the labels whereby we use sub-word units with values of 2 to the min and 5 to the max. 
To compare both \ac{KGE} types, we dubbed the KG-NMT approach that relies on \ac{EL} and Structured-based \ac{KGE} as \textit{KG-NMT (EL+KGE)}. The version with semantic information is named \textit{KG-NMT (SemKGE)}. For training, we attempted to be as generic as possible. Thus, our training set consists of a merge of the initial one-third of JRC-Acquis 3.0~\cite{steinberger2006jrc}, Europarl~\cite{koehn2005europarl} and OpenSubtitles2013~\cite{TIEDEMANN12.463}, obtaining a parallel training corpus of two million sentences, containing around 38M running words. We used the English and German versions of DBpedia as our reference \ac{KG}. The English \ac{KB} contains 4.2 million entities, 661 relations, and 2.1 million labels, while the German version has 1 million entities, 249 relations, and 0.5 million labels. As the measurement of translation quality is inherently subjective, we used three automatic \ac{MT} metrics to ensure a consistent and clear evaluation. Besides {\sc BLEU}~\cite{papineni2002bleu}, we use { \sc METEOR}~\cite{banerjee2005meteor} and {\sc chrF3}~\cite{popovic2017chrf++} on the \textit{newstest} between 2014 and 2018 for testing the models. Moreover, we carried out a manual analysis of outputs for assuring the contribution from \ac{KGE} and we investigated the use of \ac{KGE} in other settings. 

\begin{table*}[htb]
\footnotesize
\setlength\tabcolsep{1.75pt}
\centering
\begin{tabular}{@{}l|lccc|ccc|ccc|ccc|ccc@{}}
& \textbf{Models} & \multicolumn{3}{c}{\textbf{newstest2014}} & \multicolumn{3}{c}{\textbf{newstest2015}} & \multicolumn{3}{c}{\textbf{newstest2016}} & \multicolumn{3}{c}{\textbf{newstest2017}} & \multicolumn{3}{c}{\textbf{newstest2018}} \\
\toprule
& & Bleu & Met & chrF3  & Bleu & Met & chrF3 & Bleu & Met & chrF3 & BBleu & Met & chrF3 & Bleu & Met & chrF3 \\
\toprule
\multirow{3}{*}{\rotatebox{90}{Word}}
& RNN baseline & 14.47 & 33.52 & 40.03 & 16.77 & 35.20 & 41.11 & 18.55 & 36.62 & 42.54 & 15.1 & 33.75 & 39.52 & 20.53 & 39.02 & 43.92 \\
& KG-NMT (EL+KGE) & 17.19 & 36.61 & 42.14 & 19.86 & 38.25 & 42.92 & 22.38 & 40.40 & 45.18 & 18.04 & 36.94 & 41.55 & 24.87 & 43.49 & 46.88 \\
& KG-NMT (SemKGE) & \textbf{18.58} & \textbf{38.42} & \textbf{43.55} & \textbf{21.49} & \textbf{40.19} & \textbf{44.72} & \textbf{24.01} & \textbf{42.47} & \textbf{46.84} & \textbf{19.66} & \textbf{38.89} & \textbf{43.11} & \textbf{27.02} & \textbf{45.77} & \textbf{48.70} \\
\midrule
\multirow{3}{*}{\rotatebox{90}{CopyM}}
& RNN baseline& 16.75 & 37.16 & 44.93 & 19.63 & 39.20 & 46.38 & 21.37 & 40.90 & 47.85 & 17.88 & 37.89 & 44.85 & 24.22 & 43.96 & 50.15 \\
& KG-NMT (EL+KGE) & 19.53 & 39.88 & 47.18 & 22.46 & 41.67 & 48.28 & 25.05 & 44.23 & 50.66 & 20.77 & 40.58 & 47.04 & 28.44 & 47.86 & 53.25 \\
& KG-NMT (SemKGE) & \textbf{20.97} & \textbf{41.55} & \textbf{48.39} & \textbf{24.08} & \textbf{43.43} & \textbf{49.72} & \textbf{26.70} & \textbf{46.08} & \textbf{52.05} & \textbf{22.30} & \textbf{42.37} & \textbf{48.36} & \textbf{30.55} & \textbf{49.92} & \textbf{54.71} \\
\midrule
\multirow{3}{*}{\rotatebox{90}{BPE32}}
& RNN baseline & 16.33 & 38.93 & 49.82 & 15.89 & 36.51 & 45.97 & 21.95 & 42.88 & 52.68 & 16.8 & 39.12 & 49.35 & 23.85 & 45.85 & 54.98 \\
& KG-NMT (EL+KGE) & N/A & N/A & N/A & N/A & N/A & N/A & N/A & N/A & N/A & N/A & N/A & N/A & N/A & N/A & N/A \\
& KG-NMT (SemKGE) & \textbf{19.03} & \textbf{39.82} & \textbf{49.64} & \textbf{21.74} & \textbf{41.41} & \textbf{50.04} & \textbf{24.86} & \textbf{44.32} & \textbf{52.59} & \textbf{20.45} & \textbf{40.62} & \textbf{49.45} & \textbf{28.02} & \textbf{47.51} & \textbf{55.16} \\
\bottomrule
\end{tabular}
\vspace{-2mm}
\caption{Results of RNN models in BLEU (Bleu), METEOR (Met), chrF3 on WMT newstest datasets. \textit{Word} $\rightarrow$ word-based models, \textit{CopyM} $\rightarrow$ Copy Mechanism and \textit{BPE32} $\rightarrow$ BPE models.}
\vspace{0mm}
\label{tbl:newstest-birnn}
\end{table*}
\vspace{-2mm}
\paragraph{RNN vs Transformer.} Previous work has compared \ac{NN} architectures on a variety of \ac{NLP} tasks~\cite{yin2017comparative,linzen2016assessing,bernardy2017using}. However, few investigated \ac{RNN} and Transformer architectures on the translation task. Recently, \citet{tran2018importance} concluded that \ac{RNN} performs better than Transformer on a subject-verb agreement task, while \citet{tang2018evaluation} found that Transformer models surpass RNN models only in high-resource conditions. Lastly, \citet{tang2018self} compared RNN and Transformers on subject-verb agreement and \ac{WSD} by scoring contrastive translation pairs. Their findings show that Transformer models overcome RNN at \ac{WSD} task, showing that they are better at extracting semantic features. In this sense, we decided to perform a comparison between both architectures in order to analyze our hypothesis with \ac{KG}s. To build a Transformer-based KG-NMT model, we followed the specifications found at \citet{vaswani2017attention}, which use a 6-layer encoder-decoder, a batch size of 4076, 8 heads, word embeddings and hidden layers of size 512. The Adam optimizer with a learning rate of 2 and a dropout of 0.1 was used. We used the same values to sentence length, beam, and BPE. 
\vspace{-2mm}
\paragraph{Monolingual Embeddings vs. \ac{KGE}.} Here, we aim to compare the performance of an \ac{NMT} using pre-trained monolingual embeddings with the Semantically-enriched \ac{KGE} as both can be used to initialize the internal vectors' values of an \ac{NMT} model. Our focus is to analyze if the \ac{KGE} with fewer words and vectors can perform better than the monolingual embeddings for addressing the translation of entities and terminologies. We used the pre-trained monolingual embeddings from \citet{bojanowski2017enriching} for English which has 9.2 billion words and the German from \citet{grave2018learning} with 1.3 billion words. 

\section{Results}

\vspace{-2mm}
\paragraph{Overall results}

\autoref{tbl:newstest-birnn} depicts the results from KG-NMT using \ac{RNN} architecture on the \textit{newstest} dataset between 2014 and 2018. Using \ac{KGE} leads to a clear improvement over the baseline as it significantly improved the translation quality in terms of  {\sc BLEU}(+3), {\sc METEOR} (+4) and {\sc chrF3} (+3) metrics. \textit{KG-NMT (SemKGE)} outperformed \textit{KG-NMT (EL+KGE)} by around +1.3 in BLEU and chrF3, while we observe a +2 point improvement for METEOR. This difference between the contribution of \ac{KGE} types is directly related to the \ac{EL} performance which did not manage to annotate all kind of entities present in the \ac{KG}. The models on \ac{BPE} also presented consistent improvements showing that segmentation on labels of KG-NMT (SemKGE) model worked. Moreover, the use of the copy mechanism along with \ac{KGE} got the best results as expected since some entities which were not found in \ac{KG}, i.e., unfamous persons, were copied from their source words and correctly translated. For example, the entity \textit{Chad Johnston} appeared in line 1487 of the \textit{newstest}2015 dataset, but this name was not found in the \ac{KB} as an entity even though translated correctly. 

A detailed study of our results showed that the number of \ac{OOV} words decreased considerably with the augmentation through \ac{KGE}. \autoref{tab:unk} shows the number of \ac{OOV} words generated by the \ac{RNN} models across all WMT \textit{newstest} datasets. The statistics cannot ensure that every \ac{OOV} word that became a known word was essentially an entity presented in \ac{KG}. Thus, we chose the \texttt{newstest2015} for a manual analysis. First, we leveraged the {\sc METEOR} scores to identify sentences with a large number of \ac{OOV} words. We observed that many \ac{OOV} words were in fact entities contained in the \ac{KG}. As an example (line 1265), \textit{UK} was not translated by RNN baseline even using the copy mechanism (\textit{UK}) and \ac{BPE} (\textit{Britische}). However, it was correctly translated into German as \textit{Gro{\ss}britannien} by both \ac{KGE} models. Similarly, the entity \textit{Coastguard} (line 1540) was not translated correctly by baseline models, whereby both \ac{KGE} models were able to translate it into \textit{K\"{u}stenwache}. However, we observed translation mistakes regarding gender information in German. For example, while \textit{KG-NMT (EL+KGE)} was able to translate the word \textit{principal} (line 438) correctly into \textit{Direktor} but using the feminine gender (\textit{die Direktorin}). An interesting observation regarding the use of \ac{EL} is that some entities which 
were not annotated in the source text, were correctly annotated with a German \ac{URI} in the translated text. This human evaluation suggests that the \ac{KG}-augmented \ac{RNN} models were able to correctly learn the translation of entities through the relations found in \ac{KGE}.

\begin{table}[t]
\small
\setlength\tabcolsep{3pt}
\begin{tabular}{l|ccccc}
\toprule
\textbf{Model} & \textbf{2014} & \textbf{2015} & \textbf{2016} & \textbf{2017} & \textbf{2018} \\
\midrule
RNN baseline & 8,367 & 6,004 & 9,559 & 9,707 & 9,383 \\
\midrule
  + Monolingual & 5,438 & 3,832 & 5,669 & 5,624 & 6,055 \\
\midrule
KG-NMT (EL+KGE) & 6,109 & 4,427 & 6,524 & 6,603 & 6,914 \\
\midrule
KG-NMT (SemKGE) & 5,563 & 4,067 & 5,990 & 6,130 & 6,236 \\
\bottomrule
\end{tabular}
\vspace{-2mm}
\caption{Statistics of \ac{OOV} words with RNN on \textit{newstest} between 2014 and 2018.}
\vspace{-2mm}
\label{tab:unk}
\end{table}

\vspace{-2mm}
\paragraph{Comparison of RNN vs Transformer.}
\label{rnnvstrans}
\autoref{tbl:transformer-birnn} shows that \ac{KG} improved the \ac{RNN} models substantially while it decreased the performance of Transformer models on the \textit{newstest}2014-2018. The Transformer baseline word-based model outperformed the \ac{RNN} baseline word-based model across all testsets. However, once augmented with \ac{KG}, \ac{RNN}s surpassed the Transformers word-based models in {\sc BLEU}.\footnote{For the sake of space, we only display { \sc BLEU} results, but we also measured { \sc METEOR} and {\sc chrF3}.} While analyzing the translations manually on \textit{newstest2015}, we took the same aforementioned examples. In line 1265, the Transformer baseline was capable of translating \textit{UK} to \textit{Vereinigtes K\"{o}nigreich}. Also, in line 438, the transformer baseline model translated the word \textit{Principal} correctly to \textit{Direktor} with the correct male gender. In line 1540, the entity \textit{Coastguard} was not translated by any Transformer-based model. Our manual evaluation showed that Transformer models ignored the translations present in \ac{KG}. It led us to believe that the activation function played a key role for improving the semantic feature extraction. We concluded that \ac{ReLU} in Transformed was not capable of learning the \ac{KGE} values along with the word embeddings while LSTM in RNN was. 
Moreover, our findings support the results of recent studies comparing both architectures~\cite{tang2018self,tang2018analysis}.

\begin{table}[t]
\centering
\footnotesize
\setlength{\tabcolsep}{2.5pt}
\begin{tabular}{@{}lc|c|c|c|c@{}}
\toprule
\textbf{Models} & \textbf{2014} & \textbf{2015} & \textbf{2016} & \textbf{2017} & \textbf{2018} \\
\toprule
Word-based models & \multicolumn{5}{c}{}\\
\bottomrule
RNN baseline & 16.75 & 19.63 & 21.37 & 17.88 & 24.22 \\
Transformer baseline & \textbf{19.88} & \textbf{22.44} & \textbf{24.12} & \textbf{20.63} & \textbf{27.70} \\
\midrule
RNN + EL+KGE & \textbf{19.53} & \textbf{22.46} & \textbf{25.05} & \textbf{20.77} & \textbf{28.44} \\
Transformer + EL+KGE & 18.79 & 21.00 & 22.83 & 19.20 & 26.43 \\
\midrule
RNN + SemKGE  & \textbf{20.97} & \textbf{24.08} & \textbf{26.70} & \textbf{22.30} & \textbf{30.55} \\
Transformer + SemKGE & 19.10 & 21.31 & 23.22 & 19.90 & 26.84 \\
\bottomrule
BPE32 models & \multicolumn{5}{c}{}\\
\toprule
RNN baseline & 16.33 & 15.89 & 21.95 & 16.80 & 23.85 \\
Transformer baseline & \textbf{21.76} & \textbf{24.58} & \textbf{26.43} & \textbf{22.65} & \textbf{30.78} \\
\midrule
RNN+SemKGE & 19.03 & 21.74 & \textbf{24.86} & 20.45 & \textbf{28.02} \\
Transformer+SemKGE & \textbf{19.82} & \textbf{22.38} & 24.25 & \textbf{21.05} & 28.01 \\
\bottomrule
\end{tabular}
\vspace{-2mm}
\caption{Comparison between Transformer and RNN models in BLEU on various WMT \textit{newstest} datasets.}
\vspace{-2mm}
\label{tbl:transformer-birnn}
\end{table}

\vspace{-2mm}
\paragraph{Monolingual Embeddings vs \ac{KGE}.}

\autoref{tbl:comparison} reports no significant difference between monolingual embeddings and \ac{KGE} in terms of {\sc BLEU}, {\sc METEOR} and {\sc chrF3}.\footnote{Due space limitation, we only display 2017 and 2018.} At first glance, this finding is interesting since the monolingual embeddings contain billions of words, compared to the DBpedia \ac{KG} with 4.2 million entities. However, our manual analysis showed that the \ac{OOV} words addressed by the monolingual embeddings were not in fact entities, but common words and the entities remained unknown. As an example, the \textit{RNN+MonoE} model translated incorrectly the entity \textit{Principal} into \textit{Wichtigste}, while the \textit{KG-NMT (SemKGE)} used the knowledge documented in the \ac{KG}s.\footnote{\url{http://dbpedia.org/resource/Principal_(school)} to translate the entity correctly} Moreover, \textit{RNN+MonoE} was not able to translate the entities \textit{UK} and \textit{Coastguard}. 
Therefore, we envisage that a combination of both is promising and may lead to better results.

\begin{table}[t]
\centering
\footnotesize
\setlength{\tabcolsep}{1.5pt}
\begin{tabular}{@{}lccc|ccc@{}}
\textbf{Models} & \multicolumn{3}{c}{\textbf{newstest2017}} & \multicolumn{3}{c}{\textbf{newstest2018}} \\
\toprule
& {Bleu} & {Met} & {chrF3} & {Bleu} & {Met} & {chrF3} \\
\toprule
Word-based models & \multicolumn{6}{c}{}\\
\bottomrule
RNN+MonoE & \textbf{20.05} & \textbf{39.42} & \textbf{43.90} & \textbf{27.15} & \textbf{46.13} & \textbf{49.35} \\
KG-NMT (SemKGE) & 19.66 & 38.89 & 43.11 & 27.02 & 45.77 & 48.70 \\
\toprule
CopyM & \multicolumn{6}{c}{}\\
\bottomrule
RNN+MonoE & \textbf{22.61} & \textbf{42.87} & \textbf{49.01} & \textbf{30.77} & \textbf{50.39} & \textbf{55.41} \\
KG-NMT (SemKGE) & 22.30 & 42.37 & 48.36 & 30.55 & 49.92 & 54.71 \\
\toprule
BPE32 & \multicolumn{6}{c}{}\\
\bottomrule
RNN+MonoE & \textbf{20.93} & \textbf{41.41} & \textbf{50.33} & \textbf{28.42} & \textbf{48.00} & \textbf{55.98} \\
KG-NMT (SemKGE) & 20.45 & 40.62 & 49.45 & 28.02 & 47.51 & 55.16 \\
\bottomrule
\end{tabular}
\vspace{-2mm}
\caption{Comparison between pre-trained monolingual embeddings and \ac{KGE} (Met = METEOR).}
\vspace{-2mm}
\label{tbl:comparison}
\end{table}

\section{Conclusion}

In this paper, we introduced an augmentation methodology which relies on the use of \ac{KG}s to improve the performance of \ac{NMT} systems. We devised two strategies for incorporating \ac{KG} embeddings into \ac{NMT} models which works on word- and character-based models. Additionally, we carried out an extensive evaluation with a manual analysis which showed consistent enhancements provided by \ac{KG}s in \ac{NMT}. The overall methodology can be applied to any \ac{NMT} model since it does not modify the main \ac{NMT} model structure and also allows replacing different \ac{EL} systems. 



\bibliography{naaclhlt2019}
\bibliographystyle{acl_natbib}

\end{document}